\documentclass[10pt,twocolumn,letterpaper]{article}

\usepackage[pagenumbers]{cvpr2025} 
\definecolor{cvprblue}{rgb}{0.21,0.49,0.74}
\usepackage[pagebackref,breaklinks,colorlinks,allcolors=cvprblue]{hyperref}

\title{HumanDreamer: Generating Controllable Human-Motion Videos \\ via Decoupled Generation}

\author{
    Boyuan Wang\textsuperscript{\rm 1, 2}\footnotemark[1]~,
    Xiaofeng Wang\textsuperscript{\rm 1, 2}\footnotemark[1]~, 
    Chaojun Ni\textsuperscript{\rm 1, 3}, 
    Guosheng Zhao\textsuperscript{\rm 1, 2}, 
    Zhiqin Yang\textsuperscript{\rm 4},
    Zheng Zhu\textsuperscript{\rm 1}\footnotemark[2]~~,\\
    Muyang Zhang\textsuperscript{\rm 2},
    Yukun Zhou\textsuperscript{\rm 1},
    Xinze Chen\textsuperscript{\rm 1},
    Guan Huang\textsuperscript{\rm 1}, 
    Lihong Liu\textsuperscript{\rm 2},
    Xingang Wang\textsuperscript{\rm 2}\footnotemark[2]\\
    \textsuperscript{\rm 1}GigaAI~~
    \textsuperscript{\rm 2}Institute of Automation, Chinese Academy of Sciences\\
    \textsuperscript{\rm 3}Peking University~~
    \textsuperscript{\rm 4}The Chinese University of Hong Kong~~
    \\
}

\begin{document}
\twocolumn[{
\maketitle
\begin{center}
\centering
\resizebox{0.9\linewidth}{!}{
\includegraphics{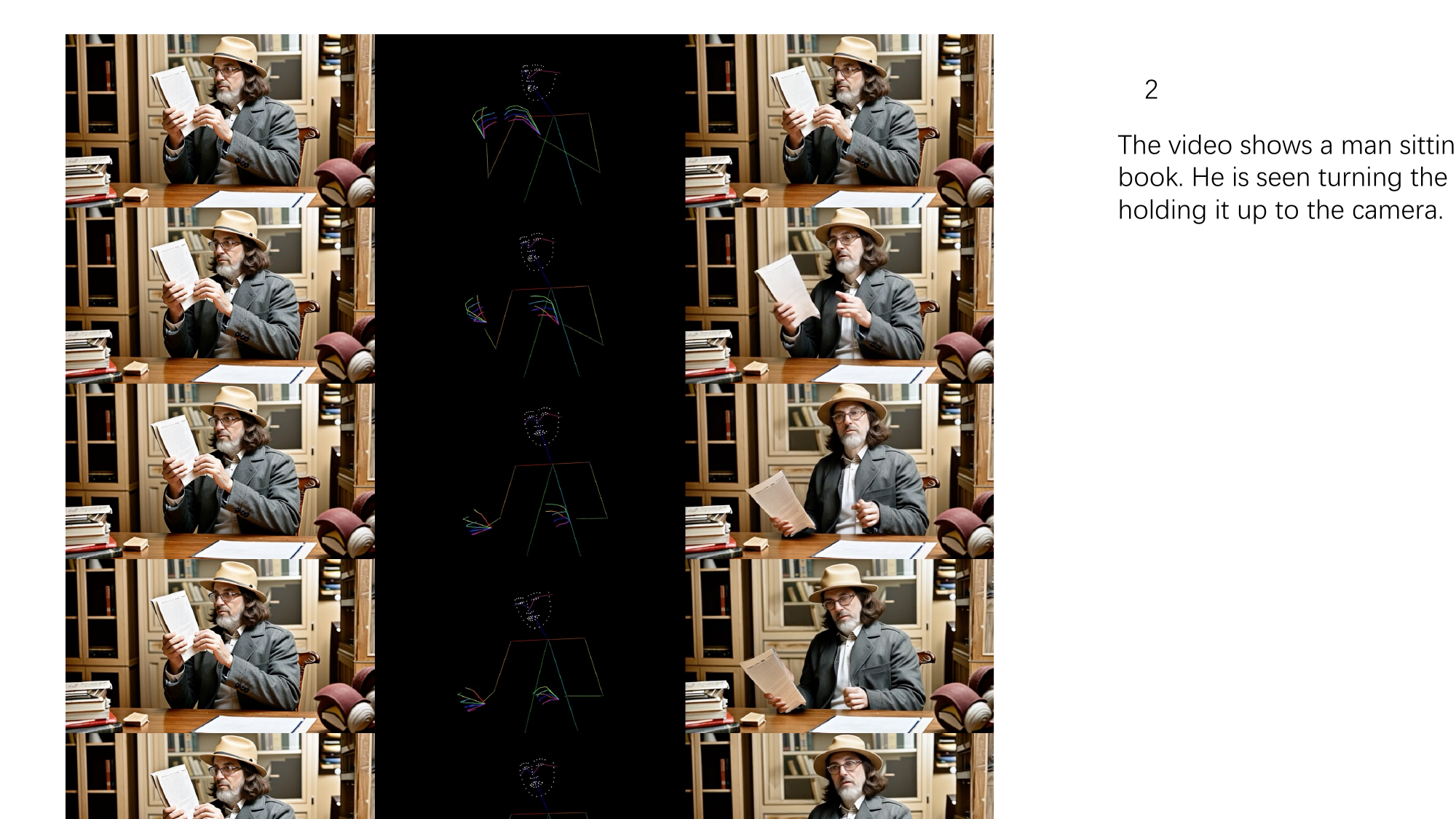}
}
\captionof{figure}{Illustration of \textit{HumanDreamer}. The human-motion video generation is decoupled into two steps: \textit{Text-to-Pose} generation and \textit{Pose-to-Video} generation. The decoupled process integrates the flexibility of text control and the controllability of pose guidance.}
\label{fig:main}
\end{center}}]

\renewcommand{\thefootnote}{\fnsymbol{footnote}}
\footnotetext[1]{
These authors contributed equally to this work. 
}
\footnotetext[2]{\mbox{Corresponding authors. zhengzhu@ieee.org, xingang.wang@ia.ac.cn.}}
\footnotetext[3]{Project Page: \url{https://humandreamer.github.io/}}

\begin{abstract}
Human-motion video generation has been a challenging task, primarily due to the difficulty inherent in learning human body movements. While some approaches have attempted to drive human-centric video generation explicitly through pose control, these methods typically rely on poses derived from existing videos, thereby lacking flexibility. To address this, we propose HumanDreamer, a decoupled human video generation framework that first generates diverse poses from text prompts and then leverages these poses to generate human-motion videos. Specifically, we propose MotionVid, the largest dataset for human-motion pose generation. Based on the dataset, we present MotionDiT, which is trained to generate structured human-motion poses from text prompts. Besides, a novel LAMA loss is introduced, which together contribute to a significant improvement in FID by 62.4\%, along with respective enhancements in R-precision for top1, top2, and top3 by 41.8\%, 26.3\%, and 18.3\%, thereby advancing both the Text-to-Pose control accuracy and FID metrics. Our experiments across various Pose-to-Video baselines demonstrate that the poses generated by our method can produce diverse and high-quality human-motion videos. Furthermore, our model can facilitate other downstream tasks, such as pose sequence prediction and 2D-3D motion lifting.
\end{abstract}

\section{Introduction}
Generating human-motion videos remains a particularly challenging task due to the inherent complexity of modeling human body movements. Despite current advancements in generative modeling \cite{zhu2024sora,guan2024world,cho2024sora,liu2024sora,sunsora}, state-of-the-art video generation models \cite{svd,hong2022cogvideo,yang2024cogvideox,pixeldance,magicvideo,walt,worlddreamer,drivedreamer4d,recondreamer}, equipped with billions of parameters and trained on millions of video and image data, still struggle to capture human body movements, frequently resulting in fragmented or unrealistic portrayals. This limitation is exacerbated when controlling human videos via textual conditions, highlighting the fundamental challenge of directly mapping text prompts to human visual data.

To enhance the generation quality, approaches such as Animate Anyone \cite{animateanyone}, UniAnimate \cite{unianimate}, Mimic Motion \cite{mimicmotion}, Champ \cite{champ} and Animate-X \cite{animatex} explicitly generate human-motion videos through pose control. By utilizing the pose-guided generation, these methods effectively reduce the complexities associated with human-motion videos. However, a notable limitation of this approach is its reliance on human-motion poses derived from existing videos, which restricts the flexibility. 

Therefore, we propose \textit{HumanDreamer}, a decoupled human-motion video generation framework that first generates human-motion poses from text prompts and subsequently produces human-motion videos based on the generated poses. The motivation behind this decoupled approach lies in that \textit{Text-to-Pose} presents a more manageable search space compared to the direct learning of \textit{text-to-pixel} representation. As shown in Fig.~\ref{fig:main}, this decoupled framework facilitates a more effective generation of human movements. Additionally, the proposed \textit{HumanDreamer} utilizes text as the input, offering greater flexibility than \cite{animateanyone,unianimate,animatex,mimicmotion,followpose} that directly rely on pre-defined poses. Specifically, we construct a 1.2 million text-pose pairs dataset \textit{MotionVid}, which is the largest dataset for human-motion pose generation. To ensure dataset quality, a comprehensive data cleaning pipeline is introduced, where the cleaning factors involve body movement amplitude, human presence duration, facial visibility, and the proportion of human figures. This rigorous cleaning guarantees that the dataset is reliable for training \textit{Text-to-Pose} tasks. Based on the dataset, we propose the \textit{MotionDiT} to generate structured human-motion poses from text prompts. The \textit{MotionDiT} integrates a global attention block designed to extract the global characteristics of the entire pose sequence, alongside a local feature aggregation mechanism that captures information from adjacent pose points, effectively combining global and local perspectives to enhance the quality and coherence of the generated poses. Additionally, the LAMA loss is utilized in the latent space to align the semantic features of the ground truth motion with those produced by \textit{MotionDiT}, which not only improves the interpretability of the model but also boosts its overall performance. The introduced techniques collectively result in a notable 62.4\% enhancement in FID, accompanied by respective increases in R-precision for the top1, top2, and top3 categories by 41.8\%, 26.3\%, and 18.3\%.

The primary contributions of this work are as follows:
(1) We present \textit{HumanDreamer}, the first decoupled framework for human-motion video generation, which integrates the flexibility of text control with the controllability of pose guidance.
(2) We propose the \textit{MotionVid}, the largest dataset for human-motion pose generation. In \textit{MotionVid}, we conduct a comprehensive data annotation and data cleaning to ensure data reliability.
(3) The \textit{MotionDiT} is introduced to generate diverse human-motion poses under text control. To enhance the generation process, we propose the LAMA loss, which significantly improves pose fidelity and diversity.
(4) Through extensive experiments, we show that \textit{MotionDiT} and the LAMA loss improve control accuracy and FID by 62.4\%, with top-k R-precision gains of 41.8\%, 26.3\%, and 18.3\% for top-1, top-2, and top-3. The generated poses also support downstream tasks like pose sequence prediction and 2D-to-3D motion enhancement.
\begin{figure*}[htbp]
    \centering
    \resizebox{\textwidth}{!}{
        \includegraphics{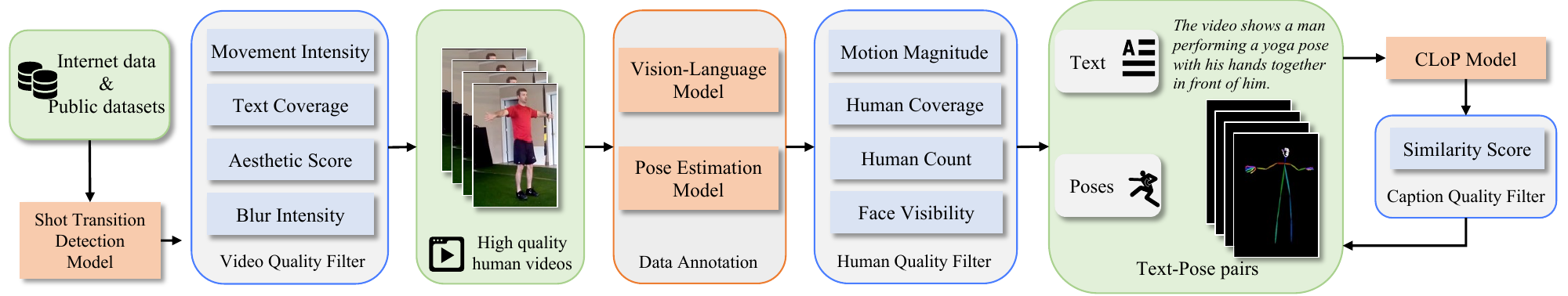}
    }
    \caption{The data cleaning and annotation pipeline for \textit{MotionVid} begins with raw data sourced from public datasets and the internet, which is then segmented into video clips. To ensure high-quality data, we apply video quality filter, data annotation, human quality filter, and caption quality filter.
}
    \label{fig:data_process}
\end{figure*}

\section{Related Work}
\subsection{Human-motion Pose Datasets}
Human motion datasets are essential for the development of human-centric perception \cite{liftedcl,nturgbd,chen2021channel,human36m,yan2018spatial,liu2019ntu} or generation \cite{motionx,motionclip,t2mgpt,mdm,diffugesture,MLD,Humantomato,Holistic,egovid} tasks. Specifically, the KIT Motion-Language Dataset \cite{kitmotion} provides sequence-level descriptions to support multi-modal motion tasks. In contrast, the HumanML3D \cite{HumanML3D} dataset, which is built upon the AMASS \cite{AMASS} and HumanAct12 \cite{Action2motion} datasets, offers richer textual annotations and a wider variety of activities. Additionally, Motion-X \cite{motionx} establishes a robust annotation protocol and is the first to compile a comprehensive fine-grained 3D whole-body motion dataset derived from extensive scene captures. As a result, existing text-driven 3D motion datasets often lack sufficient volume and diversity, which limits their scalability. A dataset that shares similarities with ours is Holistic-Motion2D \cite{Holistic}, which features 1M text-pose pairs but only 16K of these are publicly accessible. However, the proposed \textit{MotionVid} significantly surpasses this with a larger and more diverse set of 1.2M text-pose data, coupled with a more thorough data cleaning process that ensures the dataset's reliability for training \textit{Text-to-Pose} tasks.

\begin{figure*}[!t]
    \centering
    \resizebox{\textwidth}{!}{
        \includegraphics{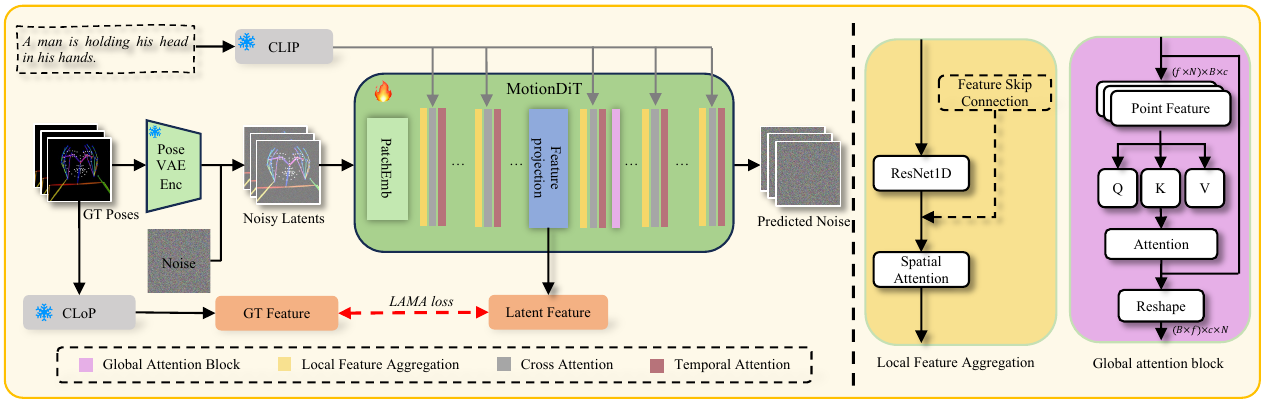}
    }
    \caption{Training pipeline of the proposed \textit{Text-to-Pose} generation. Pose data are encoded in latent space via the Pose VAE, which are then processed by the proposed \textit{MotionDiT}, where local feature aggregation and global attention are utilized to capture information from the entire pose sequence. Finally, the LAMA loss is calculated via the proposed CLoP, which enhances the training of \textit{MotionDiT}.}
    \label{fig:motiondit}
\end{figure*}

\subsection{Human-motion Pose Generation Methods}
Text-driven human motion generation, which converts textual descriptions into motion sequences, has gained significant attention in recent years. Early approaches focus on aligning motion and text through shared latent spaces. For instance, MotionCLIP \cite{motionclip} enhances autoencoder generalization by aligning the latent space with the expressive CLIP \cite{radford2021CLIP} embedding. T2M-GPT \cite{t2mgpt} frames the task as predicting discrete motion indices. More recently, Diffusion-based models, including MDM \cite{mdm}, MotionDiffuse \cite{motiondiffuse},  and DiffGesture \cite{diffugesture}, have set new performance standards. MLD \cite{MLD} introduces a latent Diffusion model that generates diverse, realistic motions, while Humantomato \cite{Humantomato} expands the task to whole-body motion generation, advancing GPT-like models for finer motion sequences. However, limited 3D data hampers model diversity and detail. Tender \cite{Holistic} overcomes this by training on extensive 2D pose data. In contrast, the proposed \textit{MotionDiT} employs a Diffusion Transformer to better capture both local and global pose relationships, improving human-motion pose generation.

\section{HumanDreamer}
In this section, we first introduce the proposed \textit{MotionVid} dataset that facilitates the training of human-motion pose generation. Next, we elaborate on the details of the decoupled \textit{Text-to-Pose} generation and \textit{Pose-to-Video} generation.
\subsection{MotionVid}

Existing datasets face limitations in training \textit{Text-to-Pose} tasks. For instance, current 3D datasets \cite{AMASS, Action2motion} primarily capture torso movements, lacking essential full-body keypoints, particularly for the face and hands, which makes them unsuitable for diverse pose generation. Although the Motion-X dataset \cite{motionx} provides full-body 3D data, its scale is limited to only 81.1K samples, as the expensive annotation of 3D data. For the 2D full-body dataset, the Holistic-Motion2D dataset \cite{Holistic} includes approximately 1M samples, though only 16K of these are publicly accessible. 

Therefore, there is a pressing need for a comprehensive and publicly available human-motion pose dataset that facilitates the training of \textit{Text-to-Pose} tasks.
To address these limitations, we introduce the largest human-motion pose dataset \textit{MotionVid}, which contains approximately 1.2M data pairs, with videos of varying resolutions and durations exceeding 64 frames. Besides, we provide a robust data collection pipeline to obtain high-quality text-pose pairs from diverse and complex data sources.

\noindent
\textbf{Data Source.} We construct a diverse set of text-pose pairs by curating data from various publicly available video datasets, including action recognition datasets Kinetics-400 \cite{kay2017kineticshumanactionvideo}, Kinetics-700 \cite{carreira2022shortnotekinetics700human}, ActivityNet-200 \cite{caba2015activitynet}, and Something-Something V2 \cite{goyal2017something}, as well as human-related datasets such as CAER \cite{lee2019context}, DFEW \cite{jiang2020dfew}, UBody \cite{lin2023osx}, Charades \cite{charades2016}, Charades-Ego \cite{sigurdsson2018charadesego}, HAA500 \cite{haa500}, and HMDB51 \cite{HMDB51}. This selection totals  $\sim$6.6M samples. To further enhance scene and subject diversity, we add 3.4M videos sourced from the internet. Given the complexity of these videos, we perform extensive annotation and cleaning to ensure high-quality data. As shown in Fig.~\ref{fig:data_process}, we first use a shot transition detection model \cite{soucek2020transnetv2} to segment the video clips, followed by the video quality filter, data annotation, human quality filter, and caption quality filter.

\noindent
\textbf{Video Quality Filter.}
After acquiring a large number of videos, it is necessary to verify the video quality. Drawing inspiration from SVD's data cleaning strategy \cite{svd}, we propose a Video Quality Filter to select high-quality video clips. 
Specifically, we use the GMFlow method \cite{xu2022gmflow} to estimate optical flow and filter out videos with insufficient movement intensity. 
Subsequently, we employ the method described in \cite{baek2019character} to detect text regions and remove videos where text occupies a significant portion of the frame. 
We also utilize LAION-AI’s aesthetic predictor \cite{aesthetics} to compute aesthetic scores and eliminate videos with low aesthetic quality.
Finally, the Laplacian operator \cite{2016Blur} is applied to assess blur intensity, and videos deemed excessively blurry are discarded. As a result, approximately 50\% of the data is filtered out (see supplementary materials for more details).

\noindent
\textbf{Data Annotation.}
We first prompt the Vision Language Model (VLM) \cite{chen2024sharegpt4video} to generate action-oriented captions that describe human activities. Besides, for each frame, we extract 2D poses using the DWPose model \cite{yang2023effective}. This process yields a large number of text-pose pairs. However, the annotated data pairs are unsuitable for training \textit{Text-to-Pose} tasks. This is due to several factors: the VLM's potential inaccuracy in describing multi-person actions, difficulty in recognizing poses occupying smaller areas in the frame, bias toward static poses from individuals with minimal motion, and limited detail in poses where faces are not visible.

\noindent
\textbf{Human Quality Filter.}
To further refine the dataset, we implement a Human Quality Filter, focusing on selecting data based on human-related characteristics. We calculate the difference between two consecutive frames' 2D poses to filter out sequences with insufficient motion magnitude. Additionally, we compute the ratio of the human detection bounding box to the entire frame to remove clips where the human coverage is too small. At this stage, we focus on simpler scenarios by selecting scenes with a human count of one and ensuring face visibility for training. Based on the Human Quality Filter, approximately 75\% of the data is filtered out (see supplementary materials for more details).

\noindent
\textbf{Caption Quality Filter.}
To enhance the quality of caption annotations, we apply a Caption Quality Filter to refine alignment accuracy. Specifically, using the proposed CLoP (see Sec.~\ref{sec:t2p} for details), we compute the semantic similarity between text and poses, filtering data based on the similarity score. This ensures that each text annotation semantically aligns with the corresponding 2D pose data, thereby improving overall data quality.

\subsection{Text-to-Pose} 
\label{sec:t2p}
Fig.~\ref{fig:motiondit} illustrates the training pipeline of the proposed \textit{Text-to-Pose} generation. In this pipeline, pose data are first encoded into a latent space via the Pose VAE and subsequently processed by the proposed \textit{MotionDiT} model. Within \textit{MotionDiT}, local feature aggregation and global attention mechanisms are employed to capture comprehensive information across the entire pose sequence. Finally, LAMA loss is computed using the proposed CLoP, which further refines the training of \textit{MotionDiT}. In the next, we introduce the Pose VAE, \textit{MotionDiT} and CLoP.

\noindent
\textbf{Pose VAE.}
Variational Autoencoders (VAEs) \cite{kingma2013auto, hazami2022efficient, cvvae, yang2024cogvideox} have been widely applied in 2D image and 3D video processing, but their use for pose data remains underexplored, presenting a rich area for research. We represent a pose sequence as \( \mathbf{p} \in \mathbb{R}^{f \times N \times 3} \), where \( f \) is the frame count, \( N \) the keypoints, and dimension 3 includes the \( x \), \( y \) coordinates and confidence scores. Given the sequential structure and frame-to-frame correlations, a 1D convolutional encoder with multi-layer downsampling effectively extracts temporal features, while symmetric decoder-encoder architecture with skip connections aids reconstruction. Confidence scores of each keypoint help mitigate occlusion issues, optimizing the model via KL divergence and reconstruction loss. Further details are available in the supplementary materials.

\noindent
\textbf{MotionDiT.}
\label{content:motiondit}
The proposed \textit{MotionDiT} extends the Diffusion Transformer (DiT) \cite{dit} architecture specifically tailored to construct associations between human-motion poses and text control.
In the proposed \textit{MotionDiT}, we incorporate a global attention block to capture global spatial-temporal patterns inherent in human poses. Besides, the \textit{MotionDiT} involves a local feature aggregation module to strengthen correlations between adjacent joints.
Specifically, we first obtain the latent representation from the Pose VAE, which is then processed via a patch-embedding:
\begin{equation}
    l_p = \mathcal{F}_\text{Patch}(\mathcal{E}(\mathbf{p})),
\end{equation}
where $\mathcal{F}_\text{Patch}(\cdot)$ is the patch embedding operation \cite{ma2024latte}, and $\mathcal{E}(\cdot)$ is the Pose VAE encoder. Then we employ Diffusion blocks to process $l_p$. As shown in Fig.~\ref{fig:motiondit}, in each block, the local feature aggregation module is first utilized to strengthen local feature correlation:
\begin{equation}
    l_p=\mathcal{F}_\text{sa}(\mathcal{F}_\text{res}(l_p)+l_p),
\end{equation}
where $\mathcal{F}_\text{res}$ is the 1D ResNet Block with kernel size 3, and $\mathcal{F}_\text{sa}$ is the spatial self attention. For text-based control, textual features obtained from CLIP \cite{radford2021CLIP} are then integrated into the network through cross-attention. Besides, following conventional video Diffusion architecture \cite{ma2024latte,svd}, temporal attention is used to ensure continuity of pose features. 
Notably, it is essential for the model to capture global pose feature information. Although the model already employs spatiotemporal attention mechanisms, certain pose characteristics at one location and time may influence other locations at different times. Thus, global attention is required to extract these global features among internal pose keypoints. Instead of incorporating global attention in every attention module, we decide to apply it to the output of the central layer of the network to reduce computational complexity. Specifically, we reshape the latent output of the middle layer to \( z \in \mathbb{R}^{(f \times n) \times c} \), where $ n $ is the number of points $ N $ divided by the down-sampling factor $ r $ and $ c $ is the number of latent channels. We then perform self-attention across all frames \( f \) and all points \( n \). Finally, we reshape the output back to its original dimensions.

\begin{table*}[!t]
    \centering
        \caption{Comparison to Other State-of-the-Art Methods on the \textit{MotionVid} Subset. The metrics demonstrate that our method outperforms others in terms of pose-text alignment and diversity. \textbf{Bold} indicates the best result.}
    \begin{tabular}{lccccccc}
        \toprule
        Method & FID $\downarrow$ & Rp-top1 $\uparrow$ & Rp-top2 $\uparrow$ & Rp-top3 $\uparrow$ & Diversity $\uparrow$  & MM Dist $\downarrow$ & MM $\uparrow$\\
        \midrule
        T2M-GPT \cite{t2mgpt}    & 496.907 & 0.176 & 0.299 & 0.395 & 64.000 & 42.950 & 14.480 \\
        PriorMDM \cite{priormdm} & 538.949 & 0.156 & 0.270 & 0.366 & 62.338 & 52.550 & 44.135 \\
        MLD \cite{MLD}           & 396.949 & 0.318 & 0.505 & 0.628 & 64.442 & 40.196 & 57.862\\
        Ours                    & \textbf{149.007} & \textbf{0.451} & \textbf{0.638} & \textbf{0.743} & \textbf{68.220} & \textbf{32.761} & \textbf{60.460} \\
        \bottomrule
    \end{tabular}
    \label{tab:main_exp}
\end{table*}

Following \cite{ma2024latte}, we employ the noise prediction loss to optimize \textit{MotionDiT}, thus the model output is:
\begin{equation}
\epsilon_{\text{pred}} = g_{\theta}(z_t, t, s),
\label{eq:formulaB}
\end{equation}
where \( g_{\theta} \) represents the \textit{MotionDiT} parameterized by \( \theta \), \( z_t \) is the noisy latent variable at time step \( t \), and \( s \) denotes the conditional input (e.g., textual features), we compute the noise prediction at time step \( t \). The model learns to predict the noise \( \epsilon_{\text{pred}} \) conditioned on the input. We calculate the mean squared error (MSE) loss between the predicted noise \( \epsilon_{\text{pred}} \) and the true noise \( \epsilon \) as:
\begin{equation}
\mathcal{L}_{d} = \mathbb{E}_{t, z_0, \epsilon} \left[ \| \epsilon - \epsilon_{\text{pred}} \|^2 \right],
\label{eq:MSE}
\end{equation}
where \( \mathcal{L}_{d} \) represents the loss over all time steps \( t \), true noise \( \epsilon \sim \mathcal{N}(0, I) \), and ground truth latent variable \( z_0 \). 

\noindent
\textbf{CLoP.} Additionally, prior works \cite{li2023your,yu2024REPA} emphasize the significance of intermediate feature representations, which inspire us to propose a \underline{LA}tent se\underline{M}antic \underline{A}lignment (\textbf{LAMA}) loss aimed at enhancing text-based motion control. We expect this term to provide benefits in two key ways: (1) By aligning the latent space of \textit{MotionDiT} with a large-scale pre-trained motion encoder, we can improve both fidelity and diversity of the generated 2D pose; (2) Aligning with the pre-trained pose encoder that is already aligned with text inherently strengthens the influence of text control on motion generation.
Unlike CLIP \cite{radford2021CLIP}, there are currently no readily available large-scale pre-trained models for aligning motion and text. Leveraging the \textit{MotionVid} dataset, we introduce \underline{C}ontrastive \underline{L}anguage-M\underline{o}tion \underline{P}re-training (\textbf{CLoP}), which aligns text with 2D pose. This improves the evaluation of 2D pose and latent semantic alignment. Specifically, inspired by CLIP, we employ a similar mechanism to extract textual features. The extracted text features and 2D pose data are then processed through dedicated text encoder \(\mathcal{F}_e(\cdot)\) and pose encoder \(\mathcal{F}_p(\cdot)\), respectively. The alignment of the two modalities is facilitated by optimizing a contrastive loss function \cite{radford2021CLIP}. Our architecture represents a 2D extension of the approach in \cite{petrovich23tmr}. Unlike the original model, we utilize CLIP for poses representation as \( \mathbf{p} \in \mathbb{R}^{f \times N \times 3} \) with its corresponding text \( \mathbf{e} \), capturing frame-wise keypoint coordinates and confidence scores instead of 3D features, the CLoP training objective as follows:
\begin{equation}
    \mathcal{L}_{c} = \ell_{ce}\left(\ell_2\left(\mathbf{h}_e\mathbf{h}_p^T\right);y\right)+\ell_{ce}\left(\ell_2\left(\mathbf{h}_p\mathbf{h}_e^T\right);y\right)/2,
\end{equation}
where \(\ell_2(\cdot)\) denote the \(\ell_2\) nomalization and \(\ell_{ce}(\cdot;y)\) represent the cross-entropy loss with ground truth label \(y\).  \( \mathbf{h}_e=\mathcal{F}_e(\mathbf{e})\textbf{W}_e\) is the projected text embedding, calculated by projection matrix \(\textbf{W}_e\). \(\mathbf{h}_p=\mathcal{F}_p(\mathbf{p})\textbf{W}_p\) is the projected pose embedding.

Based on the CLoP, we evaluate the dissimilarity within the feature space (e.g., MSE or cosine similarity metrics), achieving the benefits of the previously mentioned LAMA loss. Rather than using prior methods to extract structural features, we focus on the semantic features of motion, as motion's structural information is simpler while textual information is richer. We also employ a feature projection using a two-layer MLP \(g_{\omega}(\cdot)\) parameterized by \(\omega\) to project the latent feature space learned by \textit{MotionDiT}. As a result, the LAMA loss \(\mathcal{L}_f\) can be formulated:
\begin{equation}
    \mathcal{L}_f=d\left(g_{\omega}\left(\mathbf{h}^l_d\right), \mathbf{h}_p\right),
\end{equation}
where \(\mathbf{h}^l_d\) is the \(l\)-th layer latent representation of \textit{MotionDiT}, \(d(\cdot,\cdot)\) is the CLoP function.
Consequently, the overall objective  \(\mathcal{L}\) of \textit{MotionDiT} can be formulated as:
\begin{equation}
    \mathcal{L} = \mathcal{L}_d+\lambda_f\mathcal{L}_f,
\end{equation}
where \(\lambda_f\) is a hyperparameter to trade off the alignment and denoising process.

\subsection{Pose-to-Video}
\label{content:p2v}
We propose a model for \textit{Pose-to-Video} generation, capable of producing human-motion videos based on an initial frame image and a pose sequence. The model leverages \cite{yang2024cogvideox} as its backbone, maintaining consistent parameters with the original model, incorporates conditional control inspired by \cite{controlnet}, and employs data augmentation techniques from \cite{animatex} to enhance performance. Notably, all training was conducted on the \textit{MotionVid} dataset, demonstrating its versatility for both \textit{Text-to-Pose} and \textit{Pose-to-Video} tasks. It is worth noting that other \textit{Pose-to-Video} models, such as \cite{animateanyone, animatex, mimicmotion}, are also capable of handling this task. For further details on the model architecture, please refer to the supplementary materials.

\begin{table*}[!t]
    \centering
        \caption{The ablation study presents four configurations, progressively adding components of Local, Global, and LAMA loss to the original model. As we move from the initial configuration to the fully enhanced model, performance metrics consistently improve, highlighting the positive impact of each additional component.}
    \resizebox{\textwidth}{!}{
    \begin{tabular}{lccccccc}
        \toprule
        Method & FID $\downarrow$ & Rp-top1 $\uparrow$ & Rp-top2 $\uparrow$ & Rp-top3 $\uparrow$ & Diversity $\uparrow$ & MM Dist $\downarrow$  &MM $\uparrow$\\
        \midrule
        Vanilla DiT                  & 283.091 & 0.367 & 0.561 & 0.682 & 66.587 & 37.046 & 21.461 \\
        DiT+Local                    & 183.213 & 0.415 & 0.606 & 0.721 & 67.591 & 34.291 & 41.890 \\
        DiT+Local+Global             & 162.022 & 0.433 & 0.620 & 0.730 & 67.357 & 33.269 & 57.848 \\
        DiT+Local+Global+LAMA   & \textbf{149.007} & \textbf{0.451} & \textbf{0.638} & \textbf{0.743} & \textbf{68.220} & \textbf{32.761} & \textbf{60.460}\\
        \bottomrule
    \end{tabular}
    }
    \label{tab:ablation1}
\end{table*}
\section{Experiments}

\subsection{Experimental setup}

\noindent
\textbf{Dataset.} For computational efficiency, we select a representative subset of 50K samples from \textit{MotionVid}, covering diverse actions like dancing, squatting, and lifting to ensure diversity and expedite experimentation. This subset maintains result representativeness while enabling efficient testing. We also conducted scaling experiments to assess model performance across various dataset sizes. Qualitative visualization experiments utilize a dataset of 1.2M samples.

\begin{figure*}[!h]
    \centering
    \resizebox{\textwidth}{!}{
        \includegraphics{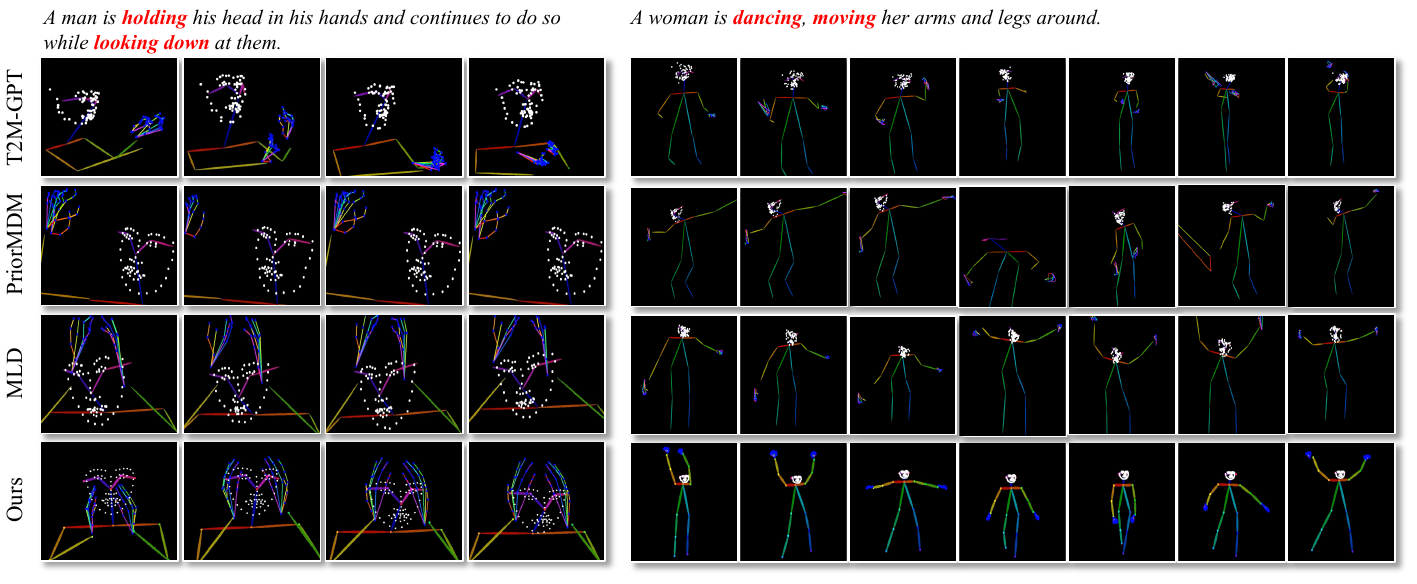}
    }
    \caption{Visualization results compared to SOTA \textit{Text-to-Pose} methods. The results demonstrate that our model significantly outperforms other models. Our method generates poses that are more consistent with the text constraints, with keypoints maintaining their integrity and minimal motion jitter. For a better visual comparison, please refer to the supplementary materials.}
    \label{fig:compare_t2p}
\end{figure*}

\noindent
\textbf{Implementation Details.}
For \textit{Text-to-Pose}, 
our motion data consists of 128 points per frame, covering the body, face, and hands. The input is structured as \( f \times 128 \times 3 \), representing \( f \) frames, with 128 points in each frame and 3 channels that encode image coordinates (x, y) along with confidence scores. To maintain uniformity across samples, all sequences are cropped to 64 frames.
For Pose VAE, we employ a downsampling factor \( r = 8 \), yielding a latent representation \( z \in \mathbb{R}^{f\times 16 \times 3} \).
In \textit{MotionDiT}, we apply a two-fold downsampling in the patch embedding. The network consists of 13 layers, with text features extracted from the SD2.1 CLIP model \cite{sd} as our text encoder, resulting in a conditional embedding \( t_{emb} \in \mathbb{R}^{1 \times 1024} \). 
The AdamW optimizer is used with a learning rate of \( 1 \times 10^{-5} \). The training is conducted on 8 H20 GPUs, with inference performed on a single H20.

\noindent
\textbf{Evaluation Metrics.}
To evaluate generated motion data quality, several metrics are employed, consistent with previous works \cite{t2mgpt,MLD,mdm,priormdm,Holistic,HumanML3D}. Frechet Inception Distance (FID) \cite{fid} assesses distributional similarity between ground truth and generated motions, using features from a custom-trained CLOP model. MultiModality evaluates diversity for a single text input, while Diversity \cite{HumanML3D} calculates variance across all motions to quantify output diversity. R-precision measures top-1, top-2, and top-3 recall rates in motion retrieval, evaluating alignment accuracy. MultiModality Distance gauges the distance between textual inputs and generated motions, indicating adherence to prompts. Further details are available in the supplementary materials.

\subsection{Results on Text-to-Pose}
Currently, Tender \cite{Holistic} is the only method for generating 2D poses based on text but its code and dataset are unavailable. Thus, we select three SOTA 3D motion generation methods for comparison: T2M-GPT \cite{zhang2023generating}, PriorMDM \cite{priormdm}, and MLD \cite{MLD}, converting their inputs to 2D poses. All methods are trained on the \textit{MotionVid} subset with proposed settings to standardize evaluation metrics. Comparison results (Tab.~\ref{tab:main_exp}) show our model outperforms others, achieving 62.4\% improvement in FID, 41.8\% in Rp-top1, 26.3\% in Rp-top2, and 18.3\% in Rp-top3. These results indicate our model generates semantically aligned outputs with high diversity.

The visual comparison with other methods is presented in Fig.~\ref{fig:compare_t2p}, and the results demonstrate a significant improvement over existing models. Our method produces poses that better align with the textual constraints, with keypoints remaining intact and exhibiting minimal motion jitter. A more detailed visual comparison can be found in the supplementary materials.

\subsection{Ablation Studies}
To validate the efficacy of the three modules proposed in the \textit{MotionDiT} (i.e., the local feature aggregation, the global attention, and the LAMA loss), we conduct a series of ablation studies. The outcomes are summarized as follows:

\noindent
\textbf{Local Feature Aggregation.} By ablating the local attention block in our model and adopting a DiT structure akin to that used in Latte \cite{ma2024latte} within each attention mechanism, we are able to demonstrate the effectiveness of our method in capturing local details. The results, presented in Tab.~\ref{tab:ablation1}, highlight the significant improvement in performance, thereby confirming the robustness of our local feature extraction strategy.

\noindent
\textbf{Global Attention Block.} To assess the impact of our global attention mechanism, we perform an ablation test where this component is removed from the model. The findings, detailed in Tab.~\ref{tab:ablation1}, indicate a notable decline in performance, which underscores the importance of the global attention mechanism in enhancing the model's ability to capture long-range dependencies and global context.

\begin{figure*}[!h]
    \centering
    \resizebox{\textwidth}{!}{
        \includegraphics{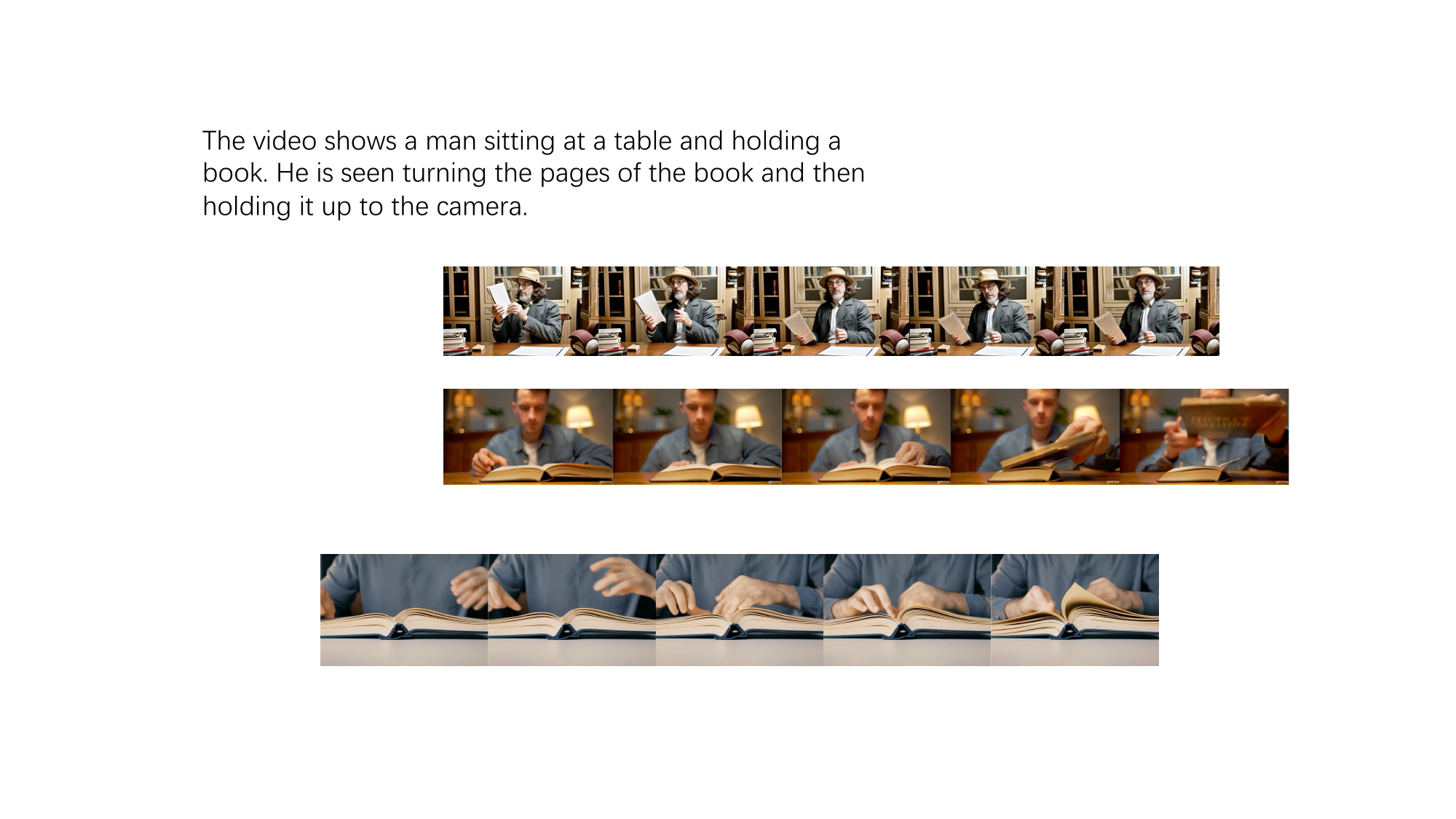}
    }
    \caption{Visualization results compared to SOTA \textit{Text-to-Video} methods. Mochi1 \cite{genmo2024mochi} and CogVideoX \cite{yang2024cogvideox} exhibit issues such as body distortion, weak motion continuity, and neglecting facial generation. In contrast, HumanDreamer is able to generate more coherent and consistent videos with smoother transitions and better attention to details such as facial expressions. For a better visual comparison, please refer to the supplementary materials.}
    \label{fig:compare_t2v}
\end{figure*}

\noindent
\textbf{LAMA Loss.} We further investigat the role of the semantic alignment loss by training \textit{MotionDiT} without it. The comparative analysis, illustrated in Tab.~\ref{tab:ablation1}, reveals a substantial reduction in the model's overall performance, thus validating the critical role of the LAMA loss in guiding the optimization process towards more accurate and stable solutions.
These ablation studies collectively provide strong evidence for the effectiveness and necessity of the proposed algorithms in improving the performance of the model across various aspects.

\subsection{Scaling Law Results}
We expand the training set size, and as shown in Tab.~\ref{tab:scale}, our metrics improved with the increase in data volume. This demonstrates the validity of using large-scale 2D pose data and suggests that model performance can be further enhanced through an efficient and cost-effective data collection pipeline.
\begin{table}[!t]
    \centering
        \caption{By increasing the amount of training data, we observe an improvement in model performance, which validates the potential for rapid scalability using large-scale 2D motion data.}
    \resizebox{\columnwidth}{!}{
    \begin{tabular}{lccccccc}
        \toprule
        Datasize  & Rp-top1 $\uparrow$ & Rp-top2 $\uparrow$ & Rp-top3 $\uparrow$ & MM Dist $\downarrow$\\
        \midrule
        50K(4\%)    & 0.451 & 0.638 & 0.743 & 32.761 \\
        250K(20\%)  & 0.452 & 0.641 & 0.751 & 32.666 \\
        500K(40\%)  & 0.464 & 0.657 & 0.766 & 32.399 \\
        1M(80\%) & 0.492 & 0.671 & 0.769 & 31.116 \\
        1.25M(100\%)& \textbf{0.513} & \textbf{0.694} & \textbf{0.791 }& \textbf{30.139} \\
        \bottomrule
    \end{tabular}
    }
    \label{tab:scale}
\end{table}

\subsection{Qualitative Results on Pose-to-Video}
In our qualitative experiments, we use the generated motion sequences as templates and provide them to the \textit{Pose-to-Video} model. The specific video results are shown in Fig.~\ref{fig:compare_t2v}. Compared to state-of-the-art text-to-video models, our model exhibits larger and more dynamic movements, with the characters' actions more accurately reflecting the textual descriptions.

\begin{figure}[htbp]
    \centering
    \resizebox{\columnwidth}{!}{
        \includegraphics{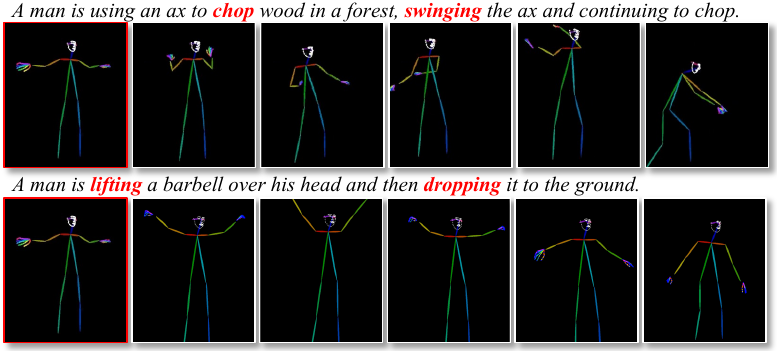}
    }
    \caption{The visualization results demonstrate that using the initial frame pose, different prompt texts can guide the generation of distinct pose sequences.}
    \label{fig:seq_pred}
\end{figure}

\subsection{Other Downstream Tasks}
In addition to the \textit{Pose-to-Video} task, the proposed \textit{Text-to-Pose} component can also be utilized for pose sequence prediction, 2D-3D motion lifting tasks.

\noindent
\textbf{Pose Sequence Prediction.} In scenarios where only partial poses sequences are available, the \textit{Text-to-Pose} model can be employed to infer and generate the missing parts of the sequence. By conditioning on both the existing pose data and a textual description of the desired movement, the model can synthesize a coherent continuation of the poses. Given the initial and final frames of a pose sequence, the model is capable of predicting the intermediate states of the poses, thereby completing the sequence, as shown in Fig.~\ref{fig:seq_pred}. This is particularly useful in applications such as animation, where incomplete poses capture data may need to be supplemented.

\begin{figure}[htbp]
    \centering
    \resizebox{\columnwidth}{!}{
        \includegraphics{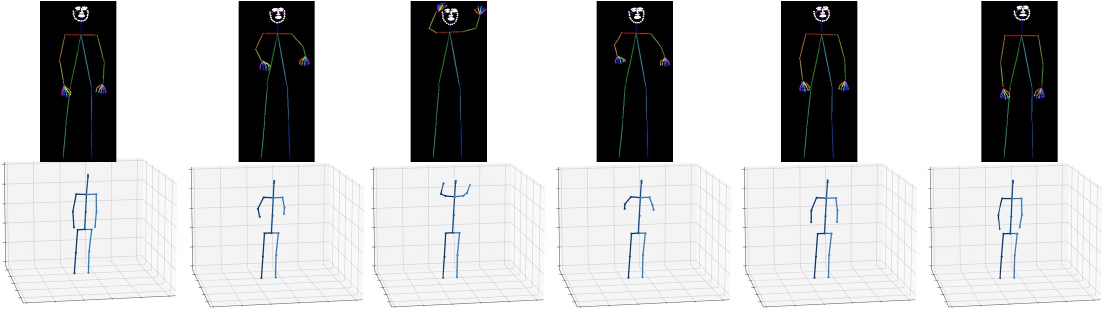}
    }
    \caption{Visualization results on 2D-3D Motion Lifting.}
    \label{fig:2d-3d}
\end{figure}

\noindent
\textbf{2D-3D Motion Lifting.}
Models like \cite{motionbert} can enhance motion data dimensionality, converting 2D motions into realistic 3D motions (Fig.~\ref{fig:2d-3d}). Given a textual description that specifies the depth or spatial aspects of the movement, the \textit{Text-to-Pose} system can generate a richer, three-dimensional representation of the motion. This capability is valuable in virtual reality (VR) and augmented reality (AR) environments, where realistic 3D motion is crucial for user experience.

\section{Discussion and Conclusion}
In this study, we present \textit{HumanDreamer}, a pioneering decoupled framework for generating human-motion videos that merges text control flexibility with pose guidance controllability. Utilizing \textit{MotionVid}, the largest dataset for human-motion pose generation, we train \textit{MotionDiT} for producing structured poses. We introduce LAMA loss to improve semantic alignment, ensuring coherent outputs.

Experimental results indicate that using generated poses in \textit{Pose-to-Video} yields high-quality, diverse human-motion videos, surpassing current benchmarks. These findings confirm the effectiveness and adaptability of our decoupled framework, facilitating versatile video generation.

\clearpage
{
    \small
    \bibliographystyle{ieeenat_fullname2025}
    \bibliography{PaperForReview}
}
\clearpage

\setcounter{section}{0}
In the supplementary material, we begin by elaborating on the implementation details of the filter and the model used in \textit{HumanDreamer}, then provide a detailed description of our proposed \textit{MotionVid} dataset, and finally present further quantitive comparison results.

\section{Implementation Details}
In this section, we detail the specific calculation methods and filtering criteria employed in the \textit{Video Quality Filter} and \textit{Human Quality Filter} within our work. Subsequently, we provide an in-depth elaboration on the implementation of PoseVAE, the pipeline of \textit{Pose-to-Video}, and the compositional specifics of the MotionVid dataset.

\subsection{Details in Video Quality Filter}
Below, we introduce the specific calculation methods and corresponding thresholds for the four filtering criteria used in the \textit{Video Quality Filter}.

\noindent
\textbf{Movement Intensity.} To assess the dynamic nature of the videos, we utilize the GMFlow method \cite{xu2022gmflow} for estimating optical flow. The purpose is to filter out videos with insufficient movement, which may not be engaging or informative. The movement intensity is defined as follows:
\begin{equation}
S_{\text{Move}} = \frac{1}{T-1} \sum_{t=1}^{T-1} \left\| \mathcal{M}(\mathbf{I}_t, \mathbf{I}_{t+1}) \right\|_{\text{avg}},
\end{equation}
where \( T \) is the total number of frames, \(\{\mathbf{I}_t\}_{t=1}^{T}\) denotes the sequence of input images over time, \(\mathcal{M}(\cdot, \cdot)\) represents the model-based optical flow prediction function, and \(\left\| \cdot \right\|_{\text{avg}}\) indicates the average magnitude of the optical flow across all pixels. The movement intensity is computed as the average of the optical flow magnitudes over consecutive frames, providing a quantitative measure of the motion within the video. Videos satisfied $ S_{\text{Move}} \leq 0.5$ are discarded to ensure that the dataset consists of content with sufficient dynamic activity.

\noindent
\textbf{Text Coverage.} To ensure the quality and readability of video content, we adopt the methodology outlined in \cite{baek2019character} for detecting text regions within frames. Following this detection, we calculate the area of each text bounding box, denoted as \( S_{\text{text}} \), and compare it against the total area of the frame, represented as \( S_{\text{frame}} \).Videos are excluded from further processing if the condition \( S_{\text{text}} > 0.07 \times S_{\text{frame}} \) is met.

\noindent
\textbf{Aesthetic Score.} To evaluate the aesthetic quality of the videos, we employ LAION-AI's aesthetic predictor \cite{aesthetics} to compute aesthetic scores. Videos with an aesthetic score $S_{\text{Aes}}$ that does not satisfy \( S_{\text{Aes}} \geq 4 \) are eliminated from the dataset. 

\noindent
\textbf{Blur Intensity.} To evaluate the sharpness of the videos, we apply the Laplacian operator \cite{2016Blur} to measure the blur intensity. The objective is to discard videos that exhibit excessive blurring, as such videos can detract from the visual quality and clarity. The blur intensity is defined as:
\begin{equation}
    S_\text{Blur} = \frac{1}{T} \sum_{i=1}^{T} \text{Var} \left( \mathcal{L} \left( \text{Gray}(\mathbf{I}_i) \right) \right),
\end{equation}
where \(\text{Gray}(\cdot)\) denotes the conversion of an RGB image to a grayscale image, \(\mathcal{L}(\cdot)\) represents the computation of the Laplacian transform, and \(\text{Var}(\cdot)\) indicates the calculation of the variance. The blur intensity \(S_\text{Blur}\) is computed as the average variance of the Laplacian-transformed grayscale images across all frames. Videos with a blur intensity \(S_\text{Blur} \leq 20\) are discarded to ensure that the dataset contains only high-quality, clear visuals.

\subsection{Details in Human Quality Filter.}
Below, we introduce the specific calculation methods and corresponding thresholds for the four filtering criteria used in the \textit{Human Quality Filter}.

\noindent
\textbf{Motion Magnitude.} To filter out sequences with insufficient motion, we calculate the difference between the 2D poses of two consecutive frames. Specifically, we compute the average difference in body keypoints between adjacent frames. The motion magnitude is defined as:
\begin{equation}
   Mag_{\text{mot}} = \frac{1}{T-1} \sum_{t=1}^{T-1} \frac{1}{N} \sum_{i=1}^{N} \left\| \mathbf{k}_i^t - \mathbf{k}_i^{t+1} \right\|,
\end{equation}
where \( N \) is the number of body keypoints, and \( \mathbf{k}_i^t \) represents the position of the \( i \)-th keypoint in the \( t \)-th frame. Videos satisfied $ Mag_{\text{mot}} \leq 10^{-3} $ are discarded to ensure that the dataset contains sequences with sufficient dynamic movement. 

\noindent \textbf{Human Coverage.} To ensure that videos contain a significant presence of human subjects, we compute the ratio of the human detection bounding box area to the entire frame area, similar to the method used for text coverage. Videos with a human coverage ratio less than $ 1/3 $ are removed from the dataset.

\noindent \textbf{Human Count.} To ensure that the videos focus on individual human subjects, we uniformly sample 5 frames from each video and count the number of detected humans in each frame. Videos are discarded if the number of humans detected in any of the sampled frames exceeds 1. 

\noindent \textbf{Face Visibility.} To ensure face visibility for training purposes, we uniformly sample 5 frames from each video. For each frame, we check the presence of 5 facial keypoints (eyes, ears, nose). If all 5 keypoints are detected in a frame, the face is considered visible. Videos are discarded if the face is not visible in any of the 5 sampled frames. 

\subsection{Details in CLoP.}
CLoP consists of two versions: one trained on a subset to filter large-scale data in \textit{Caption Quality Filter}, and another retrained on the fully filtered dataset for training and evaluating MotionDiT, similar to \cite{MLD,priormdm,t2mgpt}. CLoP is not used during MotionDiT inference.
\subsection{Details in PoseVAE.}

The pose sequence \( \mathbf{p} \in \mathbb{R}^{f \times N \times 3} \), consisting of coordinates and confidence scores, is input into a Variational Autoencoder (VAE) for reconstruction. The encoder of the VAE extracts spatial features through three layers of ResNet1D blocks and downsampling operations, which reduce spatial dimensions. This process yields a latent distribution parameterized by the mean \( \mu \) and variance \( \sigma^2 \). Using the reparameterization trick, a latent representation \( \mathbf{z} \in \mathbb{R}^{f \cdot N/8 \cdot 4} \) is sampled from this distribution. Here, \( N/8 \) reflects three rounds of downsampling, each reducing the resolution by a factor of 2, while 4 denotes the number of channels in the latent space.

The decoder reconstructs the input sequence using three layers of ResNet1D\cite{resnet1d} blocks that capture spatiotemporal features, combined with upsampling operations. This reconstruction process outputs \( \mathbf{p}_r \in \mathbb{R}^{f \times N \times 3} \). The overall architecture draws inspiration from the VAE framework proposed by \cite{svd}. 

\begin{figure}[htbp]
    \centering
    \resizebox{\linewidth}{!}{
    \includegraphics{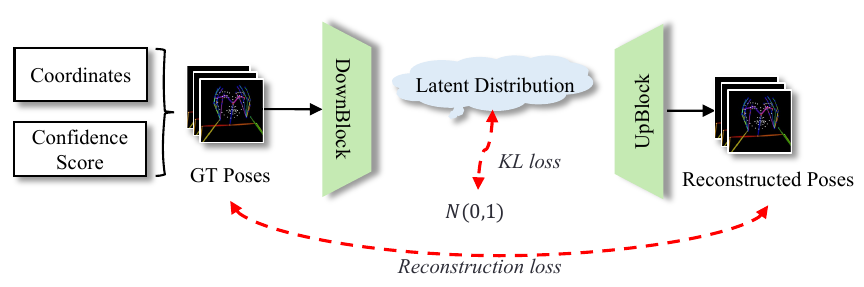}
    }
    \caption{Structure of Pose VAE.}
    \label{fig:vae}
\end{figure}

The VAE loss function, \( L_{\text{VAE}} \), consists of a reconstruction loss \( L_{\text{R}} \) and a KL divergence term \( L_{\text{KL}} \), formulated as follows:  
\begin{equation}
    L_{\text{VAE}} = L_{\text{R}} + \beta L_{\text{KL}},
\end{equation}
where \( \beta = 10^{-7} \). The reconstruction loss is defined as  
\begin{equation}
    L_{\text{R}} = \lVert \mathbf{p} - \mathbf{p}_r \rVert_2^2,
\end{equation}
and the KL divergence loss is expressed as  
\begin{equation}
    L_{\text{KL}} = \frac{1}{2} \sum_{i=1}^k \left( \sigma_i^2 + \mu_i^2 - \log(\sigma_i^2) - 1 \right),
\end{equation}
where $k$ is the dimensionality of the latent space, \( \mu \) and \( \sigma^2 \) denote the mean and variance of the latent variables' distribution, respectively. The KL divergence measures the difference between this distribution and a standard normal distribution \( \mathcal{N}(0, 1) \), which serves as the prior.
The specific architecture of the PoseVAE is illustrated in Fig.~\ref{fig:vae}.  

\subsection{Details in \textbf{\textit{Pose-to-Video}}.}
\begin{figure}[htbp]
    \centering
    \resizebox{\linewidth}{!}{
    \includegraphics{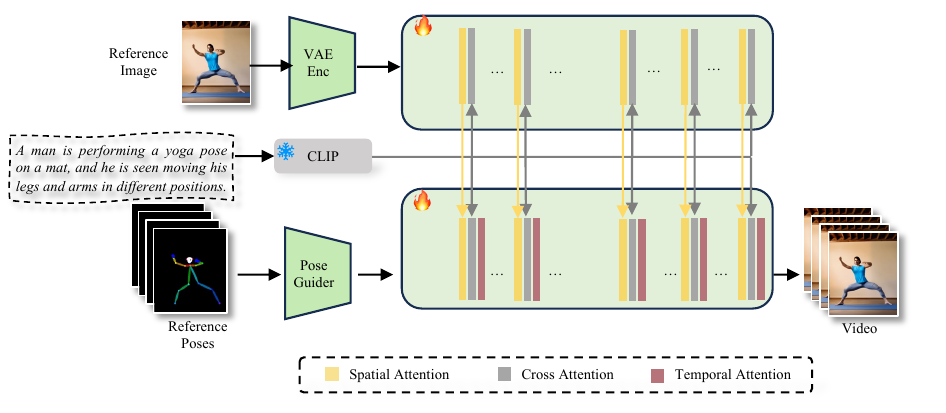}
    }
    \caption{Pipeline of \textit{Pose-to-Video}.}
    \label{fig:pose-to-video}
\end{figure}

The structure of the \textit{Pose-to-Video} model is shown in Fig.~\ref{fig:pose-to-video}. The architecture is inspired by the work in \cite{animateanyone} and utilizes the backbone from \cite{yang2024cogvideox}, which consists of stacked spatial and temporal attention layers. Textual inputs are processed through CLIP\cite{radford2021CLIP} to obtain text features, while reference poses are provided in image form to guide the generation process. The initial frame of the person can be generated from a prompt or manually specified. In our approach, we utilize SD1.5 \cite{sd} combined with ControlNet \cite{controlnet}. More advanced text-to-image models could potentially enhance alignment further. The VAE is used to encode the input conditions into a latent representation, which is then integrated into the model via cross-attention mechanisms inspired by \cite{controlnet}.

This design ensures that the generated videos are coherent and aligned with both the pose and textual inputs, leveraging advanced attention mechanisms to capture spatial and temporal dependencies effectively. 

\begin{table*}[!ht]
    \centering
    \caption{The table presents the specific composition of \textit{MotionVid}, including the sources from which it was collected, the names of the datasets, the number of clips after \textbf{video quality filter (VQF)}, the number of clips after \textbf{human quality filter (HQF)} and \textbf{caption filter (CF)}, and the data types. It shows that \textit{MotionVid} includes a diverse range of data categories, including general, action, and actions specific to different body parts, indicating a high degree of diversity.}
    \begin{tabular}{@{}p{1.4cm} p{5cm} p{2.5cm} p{2.5cm} c l@{}}
        \toprule
        \textbf{Source} & \textbf{Dataset} & \textbf{After VQF} & \textbf{After HQF+CF} & \textbf{Data Type} \\
        \midrule
        Public & Panda-70M\cite{panda70m}(partial) & 2,139,180 & 704,210 & General \\
        Public & Kinetics-700\cite{carreira2022shortnotekinetics700human} & 562,734 & 68,316 & Action \\
        Public & Kinetics-400\cite{kay2017kineticshumanactionvideo} & 298,337 & 30,855 & Action \\
        Public & Motion-X\cite{motionx} & 30,554 & 15,494 & Action \\
        Public & ActivityNet-200\cite{caba2015activitynet} & 91,220 & 7,955 & Action \\
        Public & DFEW\cite{jiang2020dfew} & 15,410 & 6,487 & Facial Action \\
        Public & CAER\cite{lee2019context} & 12,932 & 2,912 & Facial Action \\
        Public & UBody\cite{lin2023osx} & 5,981 & 2,796 & Action \\
        Public & HAA500\cite{haa500} & 8,747 & 2,133 & Action \\
        Public & HMDB51\cite{HMDB51} & 4,678 & 1,604 & Action \\
        Public & Something-Something V2\cite{goyal2017something} & 177,055 & 877 & Hand Action \\
        Public & Charades\cite{charades2016} & 10,447 & 822 & Action \\
        Public & Charades-Ego\cite{sigurdsson2018charadesego} & 8,845 & 478 & Action \\
        Internet & - & 1,686,614 & 425,382 & General \\
        \midrule
        \textbf{Total} & - & \textbf{5,052,734} & \textbf{1,270,321} & - \\
        \bottomrule
    \end{tabular}
    \label{tab:motionvid}
\end{table*}

\subsection{Details in Evaluation Metrics.}
\noindent \textbf{FID.}
In evaluating the overall quality of generated samples, the Fr\'{e}chet Inception Distance (FID) \cite{fid}is widely used. It measures the similarity between the feature distributions of real and generated data. Specifically, $\mu_{gt}$ and $\mu_{pred}$ represent the means of the feature vectors for the ground truth and predicted data, respectively, $\Sigma$ denotes the covariance matrix, and $Tr(\cdot)$ stands for the trace of a matrix. Then, FID is calculated as follows:
\begin{equation}
\text{FID} = \lVert \mu_{gt} - \mu_{pred}\rVert^2 - \text{Tr}(\Sigma_{gt} + \Sigma_{pred} - 2(\Sigma_{gt}\Sigma_{pred})^{\frac{1}{2}})
\label{eq:fid}
\end{equation}

\noindent \textbf{R-precision.}
R-precision is a metric used to evaluate the accuracy of matching between text descriptions and generated motions. It calculates the proportion of relevant items (motions) retrieved in the top-k results relative to the total number of relevant items. Specifically, it measures how many of the top-k motions correctly match their corresponding texts.

\noindent \textbf{Diversity.}
Diversity assesses the variation in motion sequences throughout the dataset. In our experiments, we randomly sample $ S_{\text{dis}} $ pairs of motions, setting $ S_{\text{dis}} $ to 300 in our experiments. Each pair's feature vectors are denoted as $ f_{\text{pred},i} $ and $ f'_{\text{pred},i} $. Diversity is then calculated by:
\begin{equation}
\text{Diversity} = \frac{1}{S_{dis}}\sum_{i=1}^{S_{dis}}||f_{pred,i} - f_{pred,i}'||
\label{formula:diversity}
\end{equation}

\noindent \textbf{MultiModality.}
MM assesses the diversity of human motions generated based on the same text description. More precisely, for the $i$-th text description, 32 motion samples are generated, and a total of 100 text descriptions are used. The features of each motion sample are extracted using CLoP. The feature vectors of the $j$-th pair derived from the $i$-th text description are represented as ($f_{\text{pred},i,j}$, $f'_{\text{pred},i,j}$). The definition of MM is given as follows:
\begin{equation}
\text{MM} = \frac{1}{32N}\sum_{i=1}^{N}\sum_{j=1}^{32}\lVert f_{pred,i,j} - f_{pred,i,j}'\rVert
\label{formula:mmodality}
\end{equation}

\noindent \textbf{MultiModality Distance.}
MM Dist measures the feature-level distance between the text embedding and the generated motion feature. The features of the i-th text-motion pair are $f_{pred,i}$ and  $f_{text,i}$. Then, MM-Dist is defined as follow:
\begin{equation}
\text{MM Dist} = \frac{1}{N}\sum_{i=1}^{N}\lVert f_{pred,i} - f_{text,i}\rVert
\label{formula:mm-dis}
\end{equation}

\section{Dataset Details}
MotionVid comprises 1.27M text-pose-video pairs, with 66.5\% originating from public datasets and 33.5\% sourced from the internet, as detailed in Tab.~\ref{tab:motionvid}. This diverse composition reflects a wide variety of styles, encompassing general, action-specific, and domain-focused clips (e.g., facial and hand actions). Notably, datasets like Panda-70M\cite{panda70m} and Kinetics-700\cite{carreira2022shortnotekinetics700human} contribute significantly to the collection, ensuring robust coverage of both general and specialized motion types. Such diversity enhances the dataset's utility for training models capable of handling heterogeneous real-world scenarios. Additionally, the inclusion of curated internet data complements the public datasets, providing more nuanced and potentially underrepresented motion patterns.

\begin{figure}[htbp]
    \centering
    \resizebox{\columnwidth}{!}{
        \includegraphics{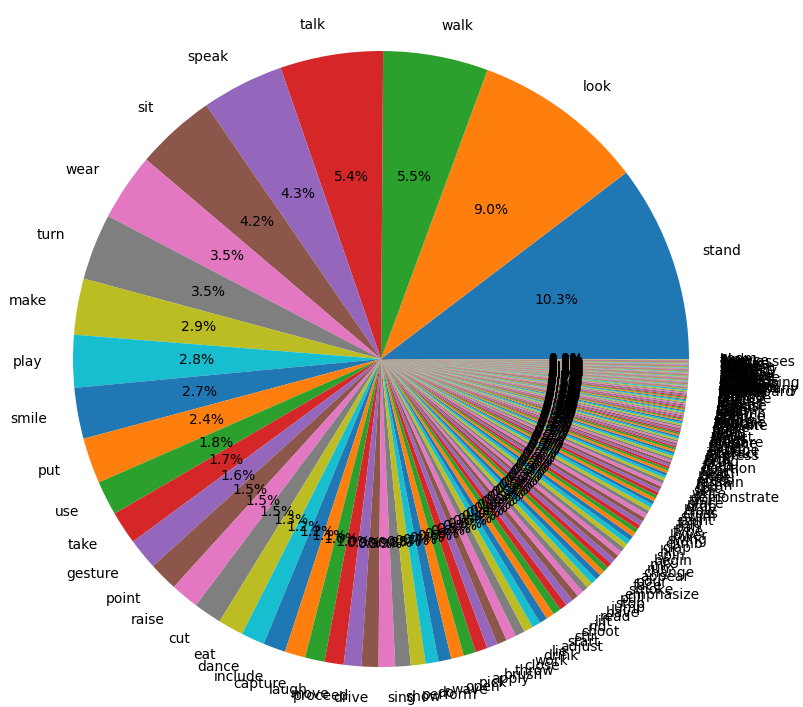}
    }
    \caption{Distribution of verbs in MotionVid's eval set.}
    \label{fig:verb_dis}
\end{figure}

The evaluation dataset, extracted from \textit{MotionVid} with 1000 samples, shows verb frequency in Fig.\ref{fig:verb_dis} after removing common verbs, indicating diverse actions. Comparisons in Tab.\ref{tab:eval_stats} show our R-precision is comparable to HumanML3D, ensuring a reasonable distribution, while \textit{Diversity} is higher, reflecting a broader range of actions and poses. The evaluation dataset’s distribution mirrors the whole dataset, which includes hundreds of action types from sources like ActivityNet200, Kinetics700, and internet data, enhancing diversity.
\begin{table}[htbp]
    \centering
    \caption{Statistics of MotionVid's eval set and HumanML3D.}
    \resizebox{\columnwidth}{!}{
    \begin{tabular}{lccccccc}
        \toprule
         Dataset & Rp-top1 $\uparrow$ & Rp-top2 $\uparrow$ & Rp-top3 $\uparrow$ & Diversity $\uparrow$\\
        \midrule
        HumanML3D  & 0.424 & 0.649 & 0.779 & 11.08  \\
        MotionVid & 0.450  & 0.639 & 0.744 & 70.11  \\
        \bottomrule
    \end{tabular}
    }
    \label{tab:eval_stats}
\end{table}
\section{Experiment Results}

Additional visualizations are presented to demonstrate the advancements in \textit{Text-to-Pose} and \textit{Pose-to-Video}, showcasing the improvements in the quality of generated videos.

\subsection{Comparison of Text-to-Pose}
We further used the poses generated by different \textit{Text-to-Pose} methods to synthesize videos, comparing the quality of the resulting human-centric videos. The results of this comparison can be found in the folder \textit{supplement/video\_in\_supplement/compare\_text\_to\_pose}, specifically in the files \texttt{Demo1.mp4} and \texttt{Demo2.mp4}.

Specifically, we employed four different models—T2M-GPT\cite{t2mgpt}, PriorMDM\cite{priormdm}, MLD\cite{MLD}, and \textit{MotionDiT}—to generate pose sequences from textual input. Subsequently, these generated poses were utilized to produce video outputs. The results indicate that our proposed method is capable of generating more stable and semantically coherent poses, which are essential for the creation of high-quality human-centric videos.

\subsection{Comparsion of Pose-to-Video}
We used the same reference image and pose sequences, but changed the models in the Pose-to-Video generation process. Specifically, we compared the video generation results using our proposed method, as well as AnimateAnyone\cite{animateanyone} and MusePose\cite{musepose}. The visualization results of this comparison can be found in the folder \textit{supplement/video\_in\_supplement/compare\_pose\_to\_video}, specifically in the files \texttt{Demo1.mp4}, \texttt{Demo2.mp4}, etc.
The results show that our proposed model achieves the best visual outcomes in video generation. 
We provide quantitative comparisons of our \textit{Pose-to-Video} with \cite{animateanyone,animatex,mimicmotion} under their experimental settings, with results summarized in Tab.~\ref{tab:p2v}. Our \textit{Pose-to-Video} demonstrates strong performance in consistency and visual quality.

\begin{table}[!t]
    \centering
        \caption{Evaluation of \textit{Pose-to-Video}.}
    \begin{tabular}{lcc}
        \toprule
        Method  & LPIPS $\downarrow$ & FVD $\downarrow$\\
        \midrule
        AnimateAnyone  & 0.285 & 171.90 \\
        MimicMotion  & 0.414 & 232.95 \\
        Animate-X   & 0.232 & 139.01 \\
        Our \textit{Pose-to-Video} & \textbf{0.148} & \textbf{116.74} \\
        \bottomrule
    \end{tabular}
    \label{tab:p2v}
\end{table}

\subsection{Comparsion of Text-to-Video}
Compared to CogVideoX, HumanDreamer excels in \textit{Sensory Quality} and \textit{Instruction Following} (CogVideoX's metrics), as confirmed by the user study on the MotionVid evaluation set (Tab.~\ref{tab:t2v}). Additionally, the \textit{Diversity} calculated from poses extracted from generated videos, shows our method outperforms CogVideoX.

\begin{table}[htbp]
    \centering
    \caption{Evaluation between HumanDreamer and CogVideoX-5B.}
    \resizebox{\columnwidth}{!}{
    \begin{tabular}{lccccccc}
        \toprule
        Method  & \shortstack{Sensory Quality $\uparrow$} & \shortstack{Instruction   Following $\uparrow$}  & Diversity  $\uparrow$\\
        \midrule
        CogVideoX-5B   & 0.531 & 0.688  & 25.285\\
        HumanDreamer  & \textbf{0.938} & \textbf{0.813}  & \textbf{68.220}\\
        \bottomrule
    \end{tabular}
    }
    \label{tab:t2v}
\end{table}


\end{document}